\documentclass{article}

\usepackage[nonatbib, final]{neurips_2019}

\usepackage[utf8]{inputenc} 
\usepackage[T1]{fontenc}    
\usepackage{hyperref}       
\usepackage{url}            
\usepackage{booktabs}       
\usepackage{amsfonts}       
\usepackage{nicefrac}       
\usepackage{microtype}      
\usepackage{amsmath}

\title{Spoken Language Understanding on the Edge}

\author{
\begin{tabular}{ccc}
~\\
 Alaa Saade & Alice Coucke & Alexandre Caulier \\
~\\
Joseph Dureau & Adrien Ball & Th\'{e}odore Bluche \\
~\\
David Leroy & Cl\'ement Doumouro & Thibault Gisselbrecht \\
~\\ 
Francesco Caltagirone & Thibaut Lavril & Ma\"{e}l Primet \\
~\\
& {\normalfont Snips, Paris, France}  \\
& {\texttt{<firstname.lastname>@snips.ai}}
\end{tabular}
}

\begin{document}

\maketitle

\begin{abstract}
  We consider the problem of performing Spoken Language Understanding (SLU) on small devices typical of IoT applications.
  Our contribution is two-fold. 
  First, we outline the design of an embedded, private-by-design SLU system and show that it has performance on-par with cloud-based commercial solutions. 
  Second, we release the datasets used in our experiments in the interest of reproducibility and in the hope that they can prove useful to the community.
\end{abstract}

\section{Introduction}
\label{sec:intro}

  Spoken Language Understanding (SLU) is the task of extracting meaning from a spoken utterance.
  Over the last years, thanks in part to steady improvements brought by deep learning approaches to Automatic Speech Recognition (ASR)~\cite{xiong2016achieving}, voice interfaces implementing SLU have greatly evolved from spotting limited and predetermined keywords to understanding arbitrary formulations of a given intention, and are becoming ubiquitous in connected devices. 
  Most current solutions however offload their processing to the cloud, where computationally demanding engines can be deployed. 
  As an example, the ASR engine achieving human parity in \cite{xiong2016achieving} is a combination of several neural networks, each containing several hundreds of millions of parameters, and large-vocabulary language models made of several millions of n-grams.
  The size of these models, along with the computational resources necessary to run them in real-time, make them unfit for deployment on small devices.
  Running SLU on the edge (i.e. embedding the engine directly on the device without resorting to the cloud) however offers several advantages. 
  First, on-device processing removes the need to send speech, or other personal data to third-party servers, therefore guaranteeing a high level of privacy. 
  In particular, we show in Section~\ref{subsec:adapated_lm} how an embedded SLU model can be personalized on device using user data. 
  Additional benefits include a reduction in latency and offline capabilities~\cite{satyanarayanan2017emergence}. 
  In this paper, we describe the Snips Voice Platform, a SLU system that runs directly on device, therefore offering all the advantages of edge computing, and has performance on-par with commercial, cloud-based solutions.

\subsection{Outline and main results}
\label{sec:outline}
  A typical SLU system has three main components. 
  First, an Acoustic Model (AM) maps a spoken utterance to a sequence of probabilities over phones (units of speech). 
  Second, a Language Model (LM) maps the output of the AM to a likely text sentence. These first two components constitue the ASR system. Third, a Natural Language Understanding (NLU) engine extracts from the sentence the intent of the user (e.g. querying the weather forecast) and the slots qualifying her query (e.g. a city in the case of a weather forecast~query). 
  Our main contribution is to outline the design of an embedded SLU system that achieves performances on-par with cloud-based solutions, and is efficient enough to run in real time on IoT devices as small as the Raspberry Pi~3, with 1GB of RAM and 1.4GHz CPU.
  This is achieved by optimizing a trade-off between accuracy and computational efficiency when designing the AM, and by contextualizing the LM and NLU components in order both to reduce their size and increase their in-domain accuracy. While the AM is trained once per language, the subsequent SLU components are use-case dependent.
  We have also released publicly\footnote{https://research.snips.ai/datasets/spoken-language-understanding} the datasets used for the experiments of Section \ref{sec:experiments} in the hope that they can be useful to the research community. 
  The NLU component of the Snips Voice Platform is open source\footnote{https://github.com/snipsco/snips-nlu}. Our SLU models can be trained through a web console, at no cost for non-commercial~use.

\subsection{Relation to previous work}
\label{sec:previous_work}
  Recent interest in mobile speech recognition has lead to new work on ASR model compression~\cite{mcgraw2016personalized}. 
  In this work, personal data is incorporated dynamically in the language model using a class-based model similar to the one we introduce in the following.
  The authors however do not study the performance of their system in terms of SLU performance but rather on a large-vocabulary speech recognition task. We rather introduce contextualized models assessed through end-to-end SLU metrics, which are arguably a better proxy for user experience \cite{wang2003word}. 
  Another line of work is interested in embedded speech commands, leveraging small models that can understand a small range of predefined commands, usually limited to one or two words \cite{warden2018speech}. 
  These approaches however cannot handle the variety of natural language interactions addressed in the following.

\section{Acoustic modeling}
\label{sec:acoustic_model}
  
  Our AM is designed so as to optimize a trade-off between accuracy and computational efficiency. 
  We use training datasets consisting of a few thousand hours of audio data with corresponding transcripts. 
  Noisy, far-field conditions with reverberation are simulated by augmenting the data with thousands of virtual rooms with random microphone and speaker locations. 
  We train deep neural AMs using the Kaldi toolkit \cite{povey2011kaldi}. Our typical architectures have $7$ layers (and one output layer), predict $\sim1600$ biphone senones, and are trained with the lattice-free Maximum Mutual Information criterion~\cite{lfmmi}, using natural gradient descent with a learning rate of $0.0005$. 
  By varying the number of neurons of each layer of the AM, we obtain models of different sizes with different computational requirements (see Table~\ref{sec:BLO}). 
  The AM is chosen to offer near state-of-the-art performance, while running in real time with acceptable memory requirements dependent on the target hardware.
  In Table~\ref{tab:net-size}, we assess the accuracy of the various architectures on a standard large-vocabulary speech recognition task with the LibriSpeech dataset \cite{panayotov2015librispeech} using the accompanying LM (refered to as \texttt{tegmed} in Kaldi).
  In the following, we consider the \texttt{nn256} model which is close to \texttt{nn512} in accuracy while being six times smaller, and runs in real time on a Raspberry Pi~3. 
  We show in the following how to compensate this loss in accuracy by contextualizing the subsequent components of the SLU pipeline to a certain domain, e.g. by restricting the vocabulary and the variety of the queries that should be modeled.

  \begin{table}
    \begin{center}
      \begin{tabular}{lcccc}
        \textbf{Layer Type} & \texttt{nn256} & \texttt{nn512} & \texttt{nn768} \\\hline
        TDNN$(-2, -1, 0, 1, 2)$  & 256 & 512 & 768 \\
        2 $\times$ TDNN$(-1, 0, 1)$ & 256 & 512 & 768 \\
        
        LSTMP(rec: -3) &  256, p128 &  512, p256 &  768, p256 \\
        2 $\times$ TDNN$(-3, 0, 3)$ & 256 & 512 & 768 \\
        
        LSTMP(rec: -3) & 256, p128 & 512, p256 & 768, p256 \\\hline
        \textbf{Num. params} & 2.6M & 8.7M & 15.4M \\\hline
      \end{tabular}
    \end{center}
    \caption{Network architecture with corresponding layer sizes. TDNN refers to a Time-Delay layer with ReLU activation. LSTMP means Long Short-Term Memory with Projection layer. A projection layer size of $N$ is denoted $pN$. The context, i.e. the number of relative frames seen by the layer at time t, is shown in parentheses: the recurrent connections skip 3 frames in LSTMP layers, and the TDNN layers consider inputs from various time steps.}
\label{sec:BLO}
  \end{table}

  \begin{table}
    \label{tab:net-size}
    \begin{center}
      \begin{tabular}{rcccc}
        \textbf{Model} & \textbf{dev-clean} & \textbf{dev-other} & \textbf{test-clean} & \textbf{test-other} \\
        \hline
        \texttt{nn256} & 7.3 & 19.2 & 7.6 & 19.6 \\
        \texttt{nn512} & 6.4 & 17.1 & 6.6 & 17.6 \\
        \texttt{nn768} & 6.4 & 16.8 & 6.6 & 17.5 \\\hline
                KALDI     & 3.9 & 10.2 & 4.2 & 10.6 \\\hline
      \end{tabular}
    \vspace{0.1cm}
    \caption{Word error rates (\%) achieved with neural networks of different sizes on the splits of the LibriSpeech dataset \cite{panayotov2015librispeech}. KALDI denotes the performance of the reference Kaldi recipe.}
    \end{center}
  \end{table}

\section{Language modeling}
\label{sec:language_model}
  
  The mapping from the output of the acoustic model to likely word sequences is done via a Viterbi search in a weighted Finite State Transducer (wFST)~\cite{mohri2001weighted}, called \emph{ASR decoding graph} in the following. 
  Formally, the decoding graph may be written as the composition of four wFSTs,
  \begin{align}
  H * C * L * G\, ,\label{eq:eager_hclg}
  \end{align}
  where $*$ denotes transducer composition, $H$ represents Hidden Markov Models (HMMs) modeling context-dependent phones, $C$ represents the context-dependency, $L$ is the lexicon and $G$ is the LM, typically a bigram or a trigram model represented as a wFST. 
  The compositions are carried out right to left, with determinization and minimization operations \cite{mohri2001weighted} applied at each step to optimize decoding. 
  We refer the interested reader to~\cite{mohri2001weighted,povey2011kaldi} and references therein for background on wFSTs and their use in speech recognition.
  In the following, we focus on the construction of the G transducer, encoding the LM, from a domain-specific dataset.

\subsection{Language model adaptation}
\label{subsec:adapated_lm}

  Our LM is adapted to understand arbitrary formulations of a finite set of intents described in a dataset. 
  Generalization to unseen queries is enabled by using both a statistical n-gram LM~\cite{katz1987estimation} which allows to mix parts of the training queries to create new ones, and class-based language modeling~\cite{brown1992class} to swap slot values. 
  More precisely, we start by building \emph{patterns} abstracting the queries of the dataset by replacing all occurrences of each slot by a symbol. 
  For example, the query ``Play some music by (The Rolling Stones)[artist]'' is abstracted to ``Play some music by ARTIST''. 
  An n-gram model is then trained on the resulting set of patterns, converted to a wFST called $G_{p}$~\cite{mohri2001weighted}. 
  Next, for each slot $s_i$ where $i\in[1, n]$ and $n$ is the number of slots, an acceptor $G_{s_i}$ is defined to encode the values the slot can take. 
  $G_{s_i}$ can either encode an n-gram model trained on a gazetteer (i.e. a list of possible values), or a generative grammar exhaustively describing the construction of any slot value (e.g. for numbers or dates). 
  Denoting wFST replacement as ``Replace'', we have~\cite{horndasch2016add}
  \begin{align}
  G = \text{Replace}(G_{p}, \{G_{s_i}\, ,\forall i\in[1, n] \})\, ,\label{eq:eager_replace}
  \end{align}
  The resulting SLU system is contextualized, and supported on a domain-specific vocabulary. 
  As a result, while a sufficient amount of specific training data may guarantee sampling the important words which allow to discriminate between different intents, it will in general prove unable to correctly sample filler words from general spoken language. 
  In order to fix this and detect out of vocabulary words (OOV), we use an approach based on so-called confusion networks \cite{xu2011minimum} to represent decoded words along with their posterior probability. We finally tag decoded words as unknown if their posterior probability is lower than some threshold.

\subsection{Dynamic language model}
\label{sec:dynamic_lm}

  On small devices, computing the decoding graph~(\ref{eq:eager_hclg}) can result in a prohibitively large wFST for larger assistants. 
  For this reason, we build a \emph{dynamic} language model by precomputing $HCL$ and G, and composing them \emph{lazily}~\cite{allauzen2010filters}. 
  The states and transitions of the  decoding graph are thus computed on demand during inference, notably speeding up the building of the LM. 
  Additionally, employing lazy composition allows to break the decoding graph into two pieces, with sizes typically much smaller than the equivalent, statically-composed HCLG. 
  When using a dynamic LM, a better composition algorithm must be used in order to keep the decoding fast enough. 
  We use composition filters~\cite{allauzen2010filters} such as look-ahead filters followed by label reachability filters with weights and labels pushing, allowing to discard inaccessible and costly decoding hypotheses early in the decoding. 
  Crucially, we ensure that the lexicon verifies the so-called C1P property (i.e. each symbol has a unique pronunciation~\cite{allauzen2009generalized}) by associating a unique symbol for each pair (word, pronunciation).
  Finally, the $\text{Replace}$ operation of Equation~(\ref{eq:eager_replace}) is performed upon loading the model from disk. 
  This allows to further break the decoding graph into smaller distinct pieces: the $HCL$ transducer mapping the output of the acoustic model to words, the query language model $G_{p}$, and the slots' language models $\{G_{s_i}\, ,\forall i\in[1, n] \}$.

  Breaking down the LM into smaller, separate parts makes it possible to efficiently update it. 
  In particular, performing on-device injection of new values in the LM becomes straightforward, enabling users to customize their embedded SLU engine. 
  For instance, if we consider an assistant dedicated to making phone calls (``call (Jane Doe)[contact]''), the user's list of contacts could be added to the values of the slot ``contact'' without this sensitive data ever leaving the device. 
  To do so, the new words and their pronunciations are first added to the $HCL$ transducer, using an embedded Grapheme to Phoneme engine (G2P) to compute the missing pronunciations. 
  The new slot values are then added to the corresponding slot wFST $G_{s_i}$ by updating the counts of the n-grams. 
  The time required for the complete slot value injection procedure ranges from a few seconds for small assistants, to a few dozen seconds for larger assistants supporting a vocabulary comprising tens of thousands of words.

\section{Natural language understanding}
\label{sec:nlu}

  The NLU component performs intent classification followed by slot filling. 
  The former is implemented with a logistic regression trained on the queries from every intent. 
  The latter consists in several linear-chain Conditional Random Fields (CRFs)~\cite{crf01}, each of them trained for a specific intent. 
  While CRFs are a standard approach for slot filling~\cite{crfinslu1}, we note that more computationally demanding approaches based on deep learning models have been recently proposed~\cite{mesnil2015using}. 
  Our experiments showed that these approaches do not yield any significant gain in accuracy in the typical training size regime of custom voice assistants (a few hundred queries).
  Data sparsity is addressed by integrating features based on precomputed word clusters, obtained by clustering word embeddings computed on a large independent corpus, effectively reducing the vocabulary size from typically 50K words to a few hundred clusters.
  Finally, gazetteer features are used, based on parsers built from the slot values provided in the training data. 
  Consistently with the n-gram slot models $G_{s_{i}}$ in the LM (see Section~\ref{subsec:adapated_lm}), these parsers can match partial slot values. 
  When injecting personal user data (see Section~\ref{subsec:adapated_lm}), these gazetteer parsers are augmented accordingly to cover the new slot values.
  This NLU component is open source and has been benchmarked and proven to be competitive against various commercial solutions \cite{coucke2018snips}.

\section{Numerical Results}
\label{sec:experiments}
  
  In this section, we present an end-to-end evaluation of both our SLU system and a cloud-based commercial solution, on two domains of increasing complexity posing different challenges.
  In the interest of reproducibility, the datasets used in the following are publicly available (see Section \ref{sec:outline}). 
  The trained SLU models can be obtained through the Snips web console at no cost for non-commercial use. 
  In our comparison with Google's cloud services, we used the service's built-in slots and features whenever possible in the interest of fairness. 
  For all experiments, we fix our threshold for OOV detection to $0.2$, the pattern transducer $G_p$ is a bigram model, while the $G_{s_{i}}$ corresponding to the gazetteer-based slots are trigrams (see Section \ref{subsec:adapated_lm} for definitions of these quantities).

  \textbf{Experimental setting.} Our datasets contain up to a few thousand text queries with their supervision, i.e. intent and slots, collected using an in-house data generation pipeline described in \cite{coucke2018snips}. 
  We then crowdsource the recording of these sentences and collect one spoken utterance for each text query in the dataset. 
  Far-field datasets are created by playing these utterances with a neutral speaker and record them using a microphone array positioned at a distance of 2 meters. 
  The aim of a SLU system is then, given one such spoken utterance, to predict the ground-true intent (intent classification) and slots. We measure the performance of both our SLU system and Google's cloud services in terms of F1-score on intent classification, and percentage of perfectly parsed utterances, such that both intent and slots are recovered.
  
        \begin{table}[h]
    \centering
    \begin{tabular}{@{}lcccccc@{}}
      \toprule
      \multicolumn{1}{c}{}       &  & \multicolumn{2}{c}{Close field} &  & \multicolumn{2}{c}{Far field} \\ \midrule
      Quantity                   &  & Snips          & Google         &  & Snips         & Google        \\ \cmidrule(r){1-1} \cmidrule(lr){3-4} \cmidrule(l){6-7}
      Intent classification (F1, \%) &  & 91.72          & 89.23           &  & 83.56          & 86.25          \\
      Perfect parsing (\%)       &  & 84.22           & 79.27           &  & 71.67          & 73.43          \\ \bottomrule
    \end{tabular}
    \vspace{0.2cm}
    \caption{`SmartLights'' assistant: end-to-end generalization performance compared with Google's Dialogflow cloud service on a 5-fold cross-validation experiment, in terms of F1-score in intent classification and percentage of perfectly parsed utterances (both intent and slots are recovered).}
    \label{tab:smart-lights-e2e-perf}
  \end{table}

  \begin{table*}[h]
    \setlength\tabcolsep{5pt} 
    \centering
    \begin{tabular}{@{}cllccccccccc@{}}
      \toprule
        & \multicolumn{1}{c}{} &  & \multicolumn{4}{c}{Close field} &  & \multicolumn{4}{c}{Far field} \\ \midrule
      Language                    & \multicolumn{1}{c}{Provider} &  & Tier 1 & Tier 2 & Tier 3 & Average &  & Tier 1 & Tier 2 & Tier 3 & Average \\ \cmidrule(r){1-2} \cmidrule(lr){4-7} \cmidrule(l){9-12}
      \multicolumn{1}{l}{English} & Snips                        &  & 71.27  & 67.73  & 67.21  & 68.73   &  & 42.08  & 39.36  & 35.58  & 39.01   \\
                                  & Google                       &  & 68.78  & 37.90  & 36.74  & 47.81   &  & 58.82  & 28.85  & 27.21  & 38.29   \\
      \multicolumn{1}{l}{French}  & Snips                        &  & 78.20  & 74.14  & 73.06  & 75.13   &  & 57.49  & 53.56  & 53.89  & 54.98   \\
                                  & Google                       &  & 61.04  & 33.51  & 32.38  & 42.31   &  & 36.24  & 15.83  & 13.47  & 21.85   \\ \bottomrule
    \end{tabular}
    \caption{Music assistants: percentage of perfectly parsed utterances of the form ``I want to listen to \#ARTIST''. The tiers are created using a ranking of 10k artists according to their stream counts on Spotify: Tier 1 corresponds to artists with rank between 1 and 1,000, tier 2 have ranking between 4,500 and 5,500 and tier 3 between 9,000 and 10,000. The Snips SLU system is trained on a complete music assistant handling several interactions with a smart speaker (see text). The results labeled ``Google'' correspond to replacing the Snips ASR component by Google's Speech Recognition API.
    }
    \label{table:music_experiment}
  \end{table*}


  \textbf{Small assistant.} We first consider a small assistant typical of smart home use cases, the ``SmartLights'' assistant, comprising $6$ intents allowing to turn on or off the light, or change its brightness or color. 
  It has a vocabulary size of approximately 400 words, and depends on three slots (room, brightness and color). 
  Table~\ref{tab:smart-lights-e2e-perf} shows that we reach an accuracy similar to a commercial, cloud-based solution. Our SLU system for this assistant has a total size of 15.1MB and runs in real time on a Raspberry~Pi~3.

  \textbf{Large assistant.} We then turn to a large and complex assistant allowing to control a smart speaker through playback control (volume control, track navigation, etc), but also play music from large libraries of artists, tracks, and albums. 
  In addition to the English version of the assistant, we also consider a French version which presents the additional difficulty of handling the pronunciations of many English words in French. 
  We compute cross-language pronunciations for these words 
  using a statistical English G2P, and then mapping their phonemes to the closest ones in the French phonology. 
  The vocabulary of the resulting English music assistant contains more than 65k words, corresponding to 178k pronunciations, while the French assistant has more than 70k words, with 390k pronunciations. 
  These assistants are the largest we consider, with a total size on disk of 80MB for the English version, and 112MB for the French version. They run in real time on a Raspberry Pi~3. We test these assistants on utterances of the form ``play some music by \#ARTIST'', where we sample ``\#ARTIST'' from a publicly available list of the most streamed artists on Spotify (released together with the dataset). 
  This experiment is representative of the difficulty of the SLU task, and additionally allows to estimate the performance of ASR systems as a function of the popularity of artists. 
  To this end, we consider two sets of experiments. In the first, we perform inference using a full Snips SLU engine and compute the fraction of correctly parsed utterances. 
  In a second experiment, we replace Snips ASR by Google's Speech Recognition API. 
  We find (see Table~\ref{table:music_experiment}) that the performance of cloud-based, general-purpose solutions such as Google's ASR decay rapidly with the ranking of the artist. 
  By contrast, our class-based approach outlined in Section \ref{subsec:adapated_lm} assigns similar weights to all artists, resulting in more robust performance even for less popular artists. 
  Additionally, in practice, our SLU system can incorporate user-specific tastes through value injection (see Section~\ref{sec:dynamic_lm}), e.g. by connecting privately to a user's favorite streaming service.

\section{Conclusion}
\label{sec:conclusion}
  SLU on the edge can achieve the accuracy of cloud-based solutions without compromising on user privacy while running in real time on small IoT devices. This is mainly done by optimizing a trade-off between accuracy and computational efficiency when designing the AM and by contextualizing the LM and NLU components.
  Future work includes further optimization to run our models on microcontrollers and leveraging local speaker identification to improve the decoding accuracy. 
   \small
\bibliography{paper}
\bibliographystyle{plain}

\end{document}